\documentclass[letterpaper, 10 pt, conference]{ieeeconf}  

\IEEEoverridecommandlockouts                              

\usepackage{graphics} 
\usepackage{epsfig} 
\usepackage{mathptmx} 
\usepackage{times} 
\usepackage{amsmath} 
\usepackage{amssymb}  
\usepackage[hidelinks]{hyperref}
\usepackage{color}
\usepackage[font=normalsize]{caption}
\usepackage{graphicx}
\usepackage{subcaption}

\title{\LARGE \bf
A Neural Network Mode for PX4 on Embedded Flight Controllers
}

\author{Sindre M. Hegre, Welf Rehberg,  Mihir Kulkarni and Kostas Alexis
\thanks{This work was supported by the Horizon Europe Grant Agreement No. 101119774.
All authors are with the Department of Engineering Cybernetics at the Norwegian University of Science and Technology, O.S. Bragstads Plass 2D, 7034, Trondheim, Norway (e-mails: sindrheg@stud.ntnu.no, {welf.rehberg, mihir.kulkarni,
    konstantinos.alexis}@ntnu.no).}}%

\begin{document}

\maketitle
\thispagestyle{empty}
\pagestyle{empty}

\begin{abstract}

This paper contributes an open-sourced implementation of a neural-network based controller framework within the PX4 stack. We develop a custom module for inference on the microcontroller while retaining all of the functionality of the PX4 autopilot. Policies trained in the Aerial Gym Simulator are converted to the TensorFlow Lite format and then built together with PX4 and flashed to the flight controller. The policies substitute the control-cascade within PX4 to offer an end-to-end position-setpoint tracking controller directly providing normalized motor RPM setpoints. Experiments conducted in simulation and the real-world show similar tracking performance. We thus provide a flight-ready pipeline for testing neural control policies in the real world. The pipeline simplifies the deployment of neural networks on embedded flight controller hardware thereby accelerating research on learning-based control. Both the Aerial Gym Simulator and the PX4 module are open-sourced at \url{https://github.com/ntnu-arl/aerial_gym_simulator} and \url{https://github.com/SindreMHegre/PX4-Autopilot-public/tree/for_paper}.
Video: \url{https://youtu.be/lY1OKz_UOqM?si=VtzL243BAY3lblTJ}.


\end{abstract}

\section{INTRODUCTION}

In recent years, there has been a surge in the use of Neural Networks (NNs) in aerial robotics for various use cases. Ranging from control to navigation in cluttered environments. While there are readily available frameworks for simulating and training control policies, the process of deploying them to real-world robots can be cumbersome and often requires customization of low-level flight controllers. A pipeline for deploying policies to Common Off-The-Shelf (COTS) Flight Controllers (FC) would drastically reduce the effort needed for researchers to test their methods in the real world. Thus accelerating research for learning-based control on aerial robots.
Running inference of NNs on an embedded controller poses many challenges. They have limited computing power and memory, so the NNs need to be small to satisfy both constraints. The minimalistic firmware on the boards with limited dependencies, does not offer the convenience of using standard libraries for inference, making it hard to build and deploy neural network modules on these boards.
We open-source our work, to provide an off-the-shelf solution for neural control in PX4 through a custom module integrated within the PX4-stack. The proposed solution leverages the broad hardware support of both the PX4-Autopilot \cite{meier_px4_2015} and the TensorFlow Lite Micro (TFLM) \cite{tflite_micro} library. Neural network training and deployment capabilities are tightly integrated with the Aerial Gym Simulator \cite{kulkarni_aerial_2025}, a GPU-accelerated simulation and rendering environment that was used to train the policies in this work.

\begin{figure}[t]
      \centering
      \includegraphics[width=1.\linewidth]{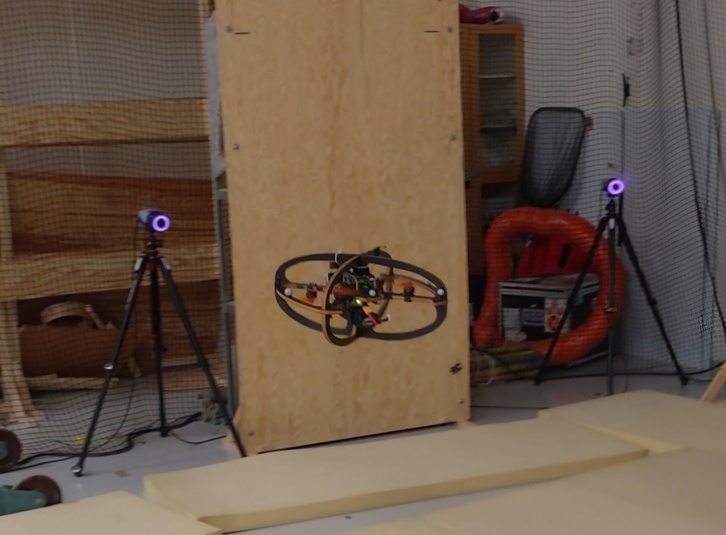}
      \caption{The Learning-based Micro Flyer (LMF) in flight with the neural controller.}
      \label{fig:lmf_controller}
\end{figure}

\section{RELATED WORKS}
Several papers have already demonstrated the capabilities of NNs on aerial robotics platforms. Previous work is often tailored to specific platforms or frameworks or requires expensive hardware. In \cite{eschmann_learning_2024}, the used firmware is specific to their platform, making it hard to leverage their method for arbitrary platforms. \cite{xing_multi-task_2024} commands collective thrusts and body-rate setpoints, leveraging the low-level control functionality provided by the FC. \cite{kulkarni_reinforcement_2024} utilizes the NVIDIA Orin NX for inference of the neural networks for high-level setpoints. \cite{bauersfeld_neurobem_2021} computes the control commands on a laptop and sends them to the onboard computer. Except for works such as in \cite{eschmann_learning_2024}, the presented efforts all use a more powerful computing device than a standard embedded FC. In \cite{shi_neural_2019}  the authors demonstrate a neural policy for just landing and takeoff using the specialized Intel Aero platform. This work indicates that our approach can be extended to support additional flight modes within the autopilot stack.
With our module being able to perform inference on the computationally constrained micro-controller of an off-the-shelf FC, we are able to deploy neural controllers that can operate at the lowest level on the auopilot without the need for external companion compute units. This further eliminates latencies introduced by the communication between high-level compute boards and FCs. The complete development environment provided with the PX4 software for validation in simulation is augmented and the support for various arbitrary multirotor platforms is added. This enables a flexible solution catering to various needs of the aerial robotics community, accelerating new solutions in this field.

\section{METHODOLOGY}
This section details the training and deployment of the neural network controllers on the embedded FC platform. To recreate this on another platform, follow the steps in the Aerial Gym documentation \url{https://ntnu-arl.github.io/aerial_gym_simulator/9_sim2real/}

\subsection{Policy Training}
In the following, the simulation framework used for this task is described. The system identification process involving the creation of a simulation model is highlighted, and the training setup is detailed.

\subsubsection{Aerial Gym Simulator}
To accurately simulate the platform, the Aerial Gym Simulator  \cite{kulkarni_aerial_2025} is utilized. The simulator is based on NVIDIA Isaac Gym~\cite{makoviychuk_isaac_2021} and provides a modular, highly parallelized simulation environment for training neural networks for arbitrary multirotor configurations. The simulator offers out-of-the-box support for control and vision-based navigation tasks. The simulator's adaptability and ease of use on new platforms enable its use in this work to train an end-to-end control policy.

\subsubsection{System Identification}\label{training}
To properly simulate the system, a URDF model and a corresponding robot configuration file were set up for the Aerial Gym Simulator. The robot's mass was measured using a weighing scale, and its inertia was estimated using a CAD model. The thrust coefficient of the robot was calculated by measuring the RPM of the motors during hovering and relating it to the force applied to each motor for steady-state hovering. Thrust-to-torque ratios were approximated, and identical motor time constants were chosen for increasing and decreasing RPM set points.

\subsubsection{RL Policy Training}
A neural-network-based position setpoint tracking controller was trained in simulation for end-to-end (e2e) control using an open-source implementation of the Proximal Policy Optimization Algorithm (PPO) provided in RL Games~\cite{makoviichuk_denys88rl_games_2025}. A reward function was established and tuned to enhance the policy's tracking performance in simulation. The reward function contained terms to reward or penalize the robot for position and orientation errors, linear and angular velocities, action magnitudes, and differences with past actions.

\subsection{Deployment}
In the following section, the core components for the network deployment will be listed. Including the autopilot software (PX4), the library for policy inference and the contributed module for PX4.

\subsubsection{PX4}

We develop this functionality within the open-source PX4 autopilot stack. PX4 \cite{meier_px4_2015} is one of the most commonly used autopilots. It supports various off-the-shelf FC hardware and has a significant adoption and an active development community. It offers a variety of safety checks and ready-to-use tools like extended Kalman filters, making it a perfect choice for use when deploying experimental NN implementations. The software is also open-sourced under the permissive BSD 3-Clause license, allowing use to the robotics community at large.

\subsubsection{TensorFlow Lite Micro}
For network inference, we rely on TensorFlow Lite for Microcontrollers (TFLM). TFLM \cite{tflite_micro} is a library for inference on embedded hardware. Its broad support for several hardware architectures, as well as its maturity, makes it a suitable solution for embedded deployment. TFLM supports a broad range of operators and there are several ways to convert networks into tflite-specific formats.

\subsection{From the Aerial Gym Simulator to PX4}
To get the trained network from the Aerial Gym Simulator and transfer into PX4, a conversion script was written as a part of this work. The script resides as a resource in the Aerial Gym Simulator repository. It converts the PyTorch network into a TFLM compatible one, and this is further converted into a C-array with a native Ubuntu command. This C-array is then copied into the PX4 module network file, ready to be deployed onto the FC.

\subsubsection{Neural Control Module in PX4}
The main contribution of this work is a neural control module, which is able to replace the classical controllers by running inference on NNs in PX4. Figure~\ref{fig:lmf_controller} shows which parts of the classical PX4 control cascade our module replaces. The current implementation provides a neural position controller that replaces the classical cascaded controllers for position, velocity, attitude, body rates, and finally, the control allocation with a single control module. The module can be easily customized to replace different selective parts of the control cascade as per the user's needs. Once included, the module starts automatically on boot and provides a new flight mode that can be chosen during flight. 
Retaining the possibility to switch to classical, stable controllers at any time facilitates testing experimental policies even on expensive hardware. Additionally, we provide a standardized testing module that publishes position set points for the controller to follow. This was used to create a simple flight path to test the performance of the controller.

\begin{figure*}[thpb]
      \centering
      \includegraphics[width=1.\linewidth]{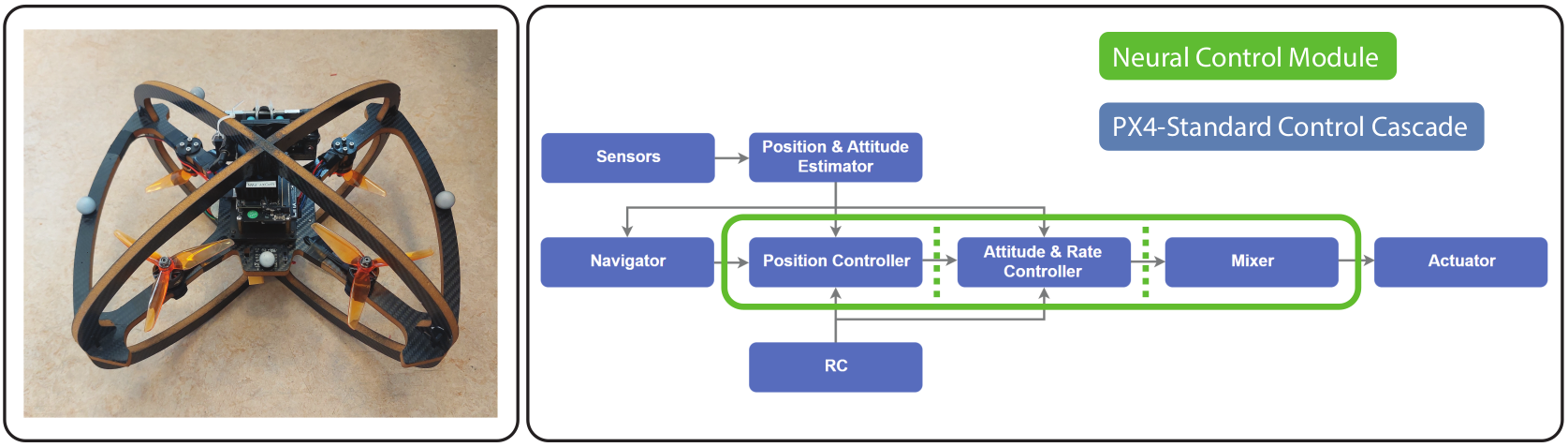}
      \caption{Left: The Learning-based Micro Flyer platform. Right: The PX4 control cascade. The NN in the Neural Control mode replaces the classical controllers as well as control allocation, which is called the mixer in the PX4 documentation. Diagram was taken from PX4 documentation \cite{px4_webpage}.}
      \label{fig:lmf_controller}
\end{figure*}

The implementation leverages the uORB middleware integrated in PX4, to retrieve the state of the drone and to publish the actuator commands to the appropriate topic. By switching which topics the module is subscribed to and publishes, all parts of the standard PX4 control cascade can be replaced by NNs. The module is scheduled as a callback every time the angular velocity state topic is received. This topic is updated at $650$ Hz on the Pixracer Pro.

The authors are currently working to make this mode a part of the official repository and offer detailed documentation. This includes more instructions and a complete overview, helping the users customize the implementation for other use cases and detailing how the different parts of the module function. In the meantime, the module can be found in the fork linked at the end of the abstract, along with instructions on using it.

\section{EXPERIMENTS}
This section describes the experimental setup and subsequent results obtained using the neural control module.

\subsection{Experimental setup}

In this work, the developed module was tested on the Learning-based Micro Flyer (LMF)~\cite{kulkarni_reinforcement_2024} from \autoref{fig:lmf_controller}, built and developed at the Autonomous Robots Lab at NTNU. It is a quadcopter in X-configuration, weighing $1.2$kg. The platform uses the mRobotics Pixracer Pro FC, which has a $32$-bit STM32H743 Cortex M7 RISC core with FPU operating at $460$ MHz, with a $2$MB flash memory and $1$MB RAM. The processor can be found on other boards as well, like the PX4 supported Pixhawk 6C, making the firmware directly applicable to other FCs. The platform also includes an NVIDIA Orin NX, which is solely used for receiving and relaying robot pose from an external motion capture unit. The flight tests have been conducted in this motion capture system to receive accurate and high-frequency pose updates.

Arming and takeoff are performed with the normal PX4 position mode for safety before switching to neural control mid-flight. The module sending position setpoints in a square pattern is subsequently initiated. The start- and end-point are situated in the middle of the square. Finally the robot controller is switched back to position mode to land the platform safely.

\subsection{Results}
\begin{figure*}[thpb]
    \centering
    \begin{minipage}{0.45\textwidth}
        \vspace{0pt} 
        \centering
        \begin{subfigure}{\textwidth}
            \includegraphics[width=\linewidth]{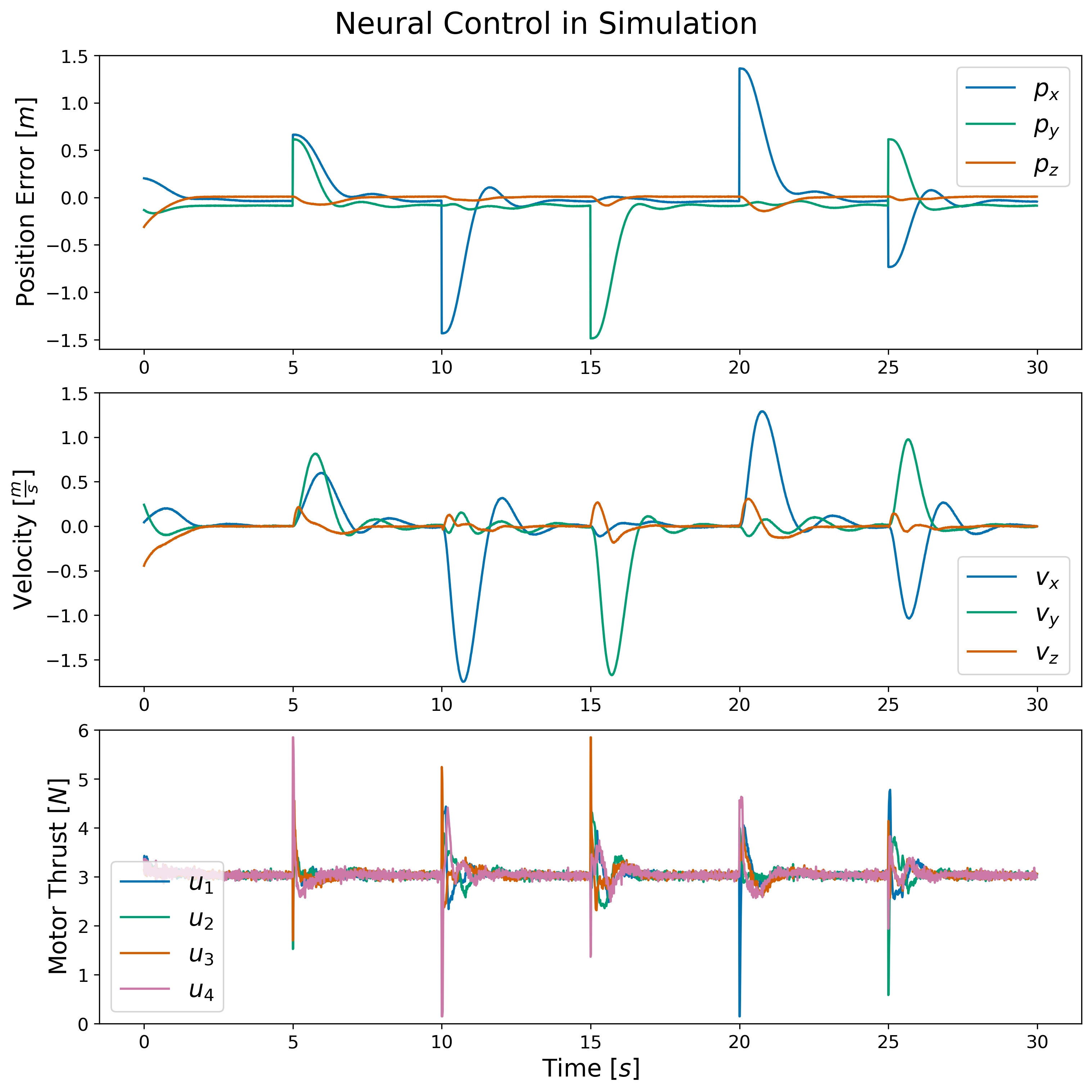}
            \label{fig:sim}
        \end{subfigure}
    \end{minipage}
    \hfill
    \begin{minipage}{0.45\textwidth}
        \vspace{0pt} 
        \centering
        \begin{subfigure}{\textwidth}
            \includegraphics[width=\linewidth]{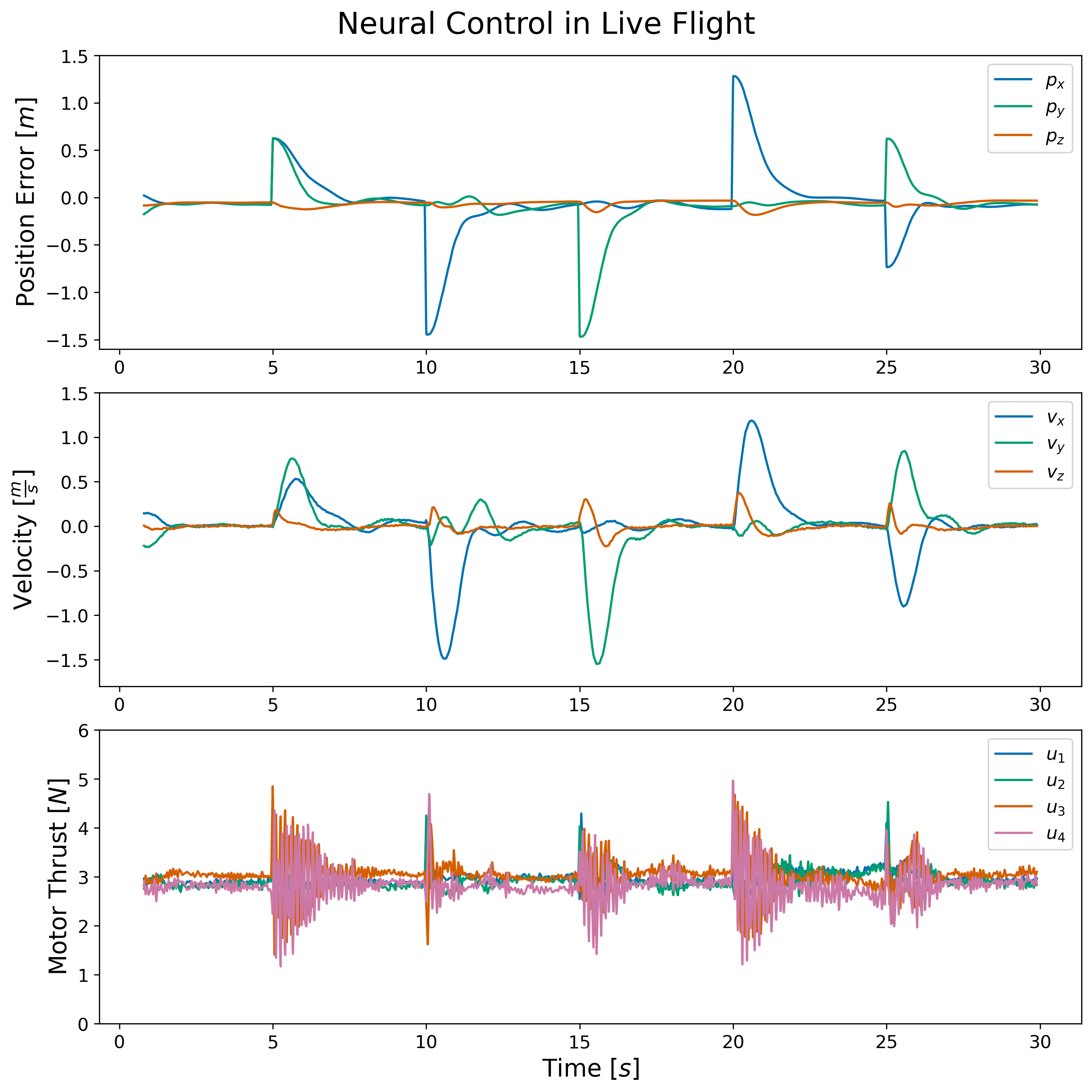}
            \label{fig:real}
        \end{subfigure}
    \end{minipage}
    \caption{Left: Performance of the neural network in simulation. Right: Performance of the neural network on the real platform.}
    \label{fig:Tests}
\end{figure*}

The NN contains 2-hidden layers with $64$ and $32$ neurons, both layers are fully connected and use ReLU as activation function. The $15$-dimensional inputs consist of observations containing state information and the $4$-dimensional outputs provide motor thrust setpoints. This network is small enough to fit within the $50$KB RAM available for use on the microcontroller board of the FC. Pose measurements from an external Qualisys motion capture setup were relayed to the flight controller and were further transformed to be represented in the relevant frames for inference. Inference requires approximately $93.4~\mu\textrm{s}$ while the entire control loop, including the pre-processing of inputs and post-processing of outputs, requires $137.6~\mu\textrm{s}$. The neural controller is shown to achieve similar performances on the real platform as it did in simulation, indicating robust sim2real transferability. The plots from the simulation and the corresponding mission in a real-world experiment are shown in Figure~\ref{fig:Tests}. A video of the flight can be seen at the end of the video linked in the abstract.

\section{DISCUSSION}

The main focus of this work is twofold: a) development of a neural control module integrated in low-level flight control software on compute- and memory-constrained micro-controllers, and b) demonstrating sim2real transfer of end-to-end motor control policies deployed on such hardware. While the response of the system in both simulation and the real-world experiment is largely consistent for the position and velocity measurements, we observe a larger variance between the motor commands across these two experiments. We believe this is caused by inaccurate estimates of the motor time constants used during training. This can be improved by measuring the time constants of the motor-propeller combination on a dedicated test bench to significantly narrow the gap. 

Having neural control as its own flight mode allows one to deploy networks that are specialized for various niche (or custom) use cases without additional consideration regarding taking off or landing safely, since the existing PX4 controllers can be engaged when necessary. Our aim with this implementation is to allow researchers and the community at large to utilize the entire autopilot software, without the need for re-implementing drivers, or setting up their own message passing middleware or recreate existing tools. The low inference time of $93.4$ $\mu$s allows the controller to run at a high frequency that may be exploited for other applications, such as control for highly-agile maneuvers. Simultaneously, direct implementation of NN-based control on FCs eliminates latencies introduced when high-level compute boards are also used. 

\subsection{Future work}
The module can be further extended to support other controller setups, such as just neural control allocation or controllers up to the off-the-shelf control allocator. Other directions of research may include the development of a neural observer. 

\section{CONCLUSION}
We have demonstrated that an NN-based controller can be deployed on an embedded flight controller as a module within the PX4 autopilot. This provides an accessible and customizable pipeline for testing neural network implementations on off-the-shelf embedded flight controllers. Open-sourced code and documentation detailing both the training setup and the inference module make the entire setup accessible and quick to deploy.



\bibliographystyle{IEEEtran}
\bibliography{root.bib}

\end{document}